# Meta-learning Based Short-Term Passenger Flow Prediction for Newly-Operated Urban Rail Transit Stations

Kuo Han[a], Jinlei Zhang[b]*, Chunqi Zhu[a], Lixing Yang[b], Xiaoyu Huang[a], Songsong Li[a]

[a] School of Civil Engineering, Beijing Jiaotong University

[b] State Key Laboratory of Rail Traffic Control and Safety, Beijing Jiaotong University

* Corresponding author: zhangjinlei@bjtu.edu.cn (J. Zhang)

# Short-Term Passenger Flow Prediction for Newly-Operated Urban Rail Transit Stations

**ABSTRACT**

Accurate short-term passenger flow prediction in urban rail transit stations has great benefits for reasonably allocating resources, easing congestion, and reducing operational risks. However, compared with data-rich stations, the passenger flow prediction in newly-operated stations is limited by passenger flow data volume, which would reduce the prediction accuracy and increase the difficulty for station management and operation. Hence, how accurately predicting passenger flow in newly-operated stations with limited data is an urgent problem to be solved. Existing passenger flow prediction approaches generally depend on sufficient data, which might be unsuitable for newly-operated stations. Therefore, we propose a meta-learning method named Meta Long Short-Term Memory Network (Meta-LSTM) to predict the passenger flow in newly-operated stations. The Meta-LSTM is to construct a framework that increases the generalization ability of long short-term memory network (LSTM) to various passenger flow characteristics by learning passenger flow characteristics from multiple data-rich stations and then applying the learned parameter to data-scarce stations by parameter initialization. The Meta-LSTM is applied to the subway network of Nanning, Hangzhou, and Beijing, China. The experiments on three real-world subway networks demonstrate the effectiveness of our proposed Meta-LSTM over several competitive baseline models. Results also show that our proposed Meta-LSTM has a good generalization ability to various passenger flow characteristics, which can provide a reference for passenger flow prediction in the stations with limited data.

**Keywords:** meta learning; short-term prediction; passenger flow; urban rail transit; newly-operated station

## 1. Introduction

Urban rail transit (URT) plays an important role in the urban transportation system. It has developed rapidly in China recently. According to the official statistical report, as of December 31, 2021, in mainland China, the URT has been operated in forty cities, with more than 200 operated lines and a total mileage of 7209.7 km. Because of its efficient and convenient characteristics, the URT has attracted a large number of passengers. The passenger flow volume has reached up to 23.69 billion in URT in 2021 despite the impact of COVID-19. In terms of such massive passenger flows, it is critically important to obtain the spatiotemporal distributions of passengers for operators providing better service. An accurate short-term passenger flow prediction for URT stations enables operators to take corresponding management measures in advance. It is beneficial for reasonably allocating resources, easing congestion, and reducing operational risks. For example, in peak and off-peak hours, the operators can adjust the train timetables and limit the passenger inflows according to the predicted results.

With the extensive application of the automatic fare collection (AFC) systems, a massive amount of historical trip data can be easily obtained and recorded (Yang et al. 2021). Meanwhile, deep learning methods have been developed rapidly. The combination of deep learning methods and massive passenger flow data makes it possible to conduct an accurate passenger flow prediction in URT stations. However, not all URT stations can provide sufficient passenger flow data in the real-world, such as newly-operated stations. The data limitation may reduce the accuracy of passenger flow prediction. It might be more difficult to conduct reasonable operation management if operators lack precise prediction results for reference, although they are experts in real-world operations. Therefore, there is a higher demand for accurate passenger flow prediction in stations with insufficient data than that for stations with abundant data.

The models used for short-term passenger flow prediction can be classified into conventional time series methods, machine learning methods, and deep learning methods. In URT and road traffic field, the conventional time series methods (e.g., historical average (HA), logistic regression, autoregressive integrated moving average (ARIMA)(Zhu 2010), Kalman filter (Jiao et al. 2016), and their variants(L. Li et al. 2018; Moreira-Matias et al. 2013) are first applied to the passenger flow prediction in the early stage. To improve the prediction accuracy, machine learning methods are applied to short-term passenger flow prediction, such as support vector machine (SVM), backpropagation neural networks (BPNNs)(P. Wang and Liu 2008a), and their hybrid models(Sun,

Leng, and Guan 2015; X. Wang et al. 2018a). With the development of computer science, deep learning methods which is a branch of machine learning are applied to short-term passenger flow prediction, such as the models based on recurrent neural networks (RNN) and convolutional neural networks (CNN)(Ma et al. 2015a, 2017, 2021; Yipeng Liu et al. 2017; Polson and Sokolov 2017). The passenger flow volume can be accurately predicted by these deep learning methods when there is large-scale data. However, the performance of the aforementioned methods might be poor when the data is insufficient.

Transfer learning methods (Ren et al. 2019) provide a solution to the problem of conducting predictions with insufficient data. It has recently been applied in road traffic, while less in URT. The transfer-learning method solves the problem by transferring the knowledge from the source domain with abundant data to the target domain with limited data (N. Li et al. 2021; Karami, Crawford, and Delp 2020). In (Wei, Zheng, and Yang 2016), to solve the problem of air quality prediction with insufficient data, the transfer learning method learns semantically related dictionaries for multiple modalities from a data-rich city and simultaneously transfers the dictionaries into a target city with insufficient data. The method proposed in (L. Wang et al. 2018) provides a solution for crowd flow prediction with limited data. They build the inter-city region pairs between the cities with adequate data and insufficient data and then calculate their similarity score to enhance the transfer performance. To address the data insufficiency problem in traffic prediction, (J. Li et al. 2021) uses data patterns and geographic attributes as criteria to select links similar to the target links (i.e., the links with insufficient data) from the links with sufficient data (i.e., source links), and transfer the knowledge to the target links. However, the performance of transferring would be unstable when the knowledge is transferred from a single source domain. For example, the performance of transferring might be favorable if the features of the source domain and target domain match well. However, if they do not match, the transfer learning methods would make no contributions or even decrease the performance (Yao et al. 2019).

As for passenger flow prediction in newly-operated URT stations, the transfer learning method would be a good option. We can transfer the knowledge from the data-rich stations to data-scarce stations. During the transferring process, to improve the stability caused by the characteristics mismatch between two domains, we consider learning the knowledge from multiple source stations and transferring the knowledge to target stations. Meta-learning (a.k.a., learning to learn)(Finn, Abbeel, and Levine 2017; Nichol, Achiam, and Schulman 2018) framework enables the model to get well generalization on a variety of learning tasks (i.e., source stations), such that it can solve new learning tasks (i.e., target stations) using only a small number of training samples. For example, in (Tian et al. 2021), the authors proposed a meta-learning method for traffic flow prediction in cities with limited data. This study extracted the knowledge of traffic flow from multiple data-rich cities and then transferred the knowledge to the cities with insufficient data, and thus the stability of transfer was increased. However, there are two major challenges to be considered when we utilize the meta-learning method for passenger flow prediction in newly-operated URT stations as follows:

**(1) Construct the meta-learning task**

A single meta-learning task contains multiple stations and the passenger flow information of these stations. We transfer the knowledge from the source stations to the target stations via meta-learning task-based learning. Yet, first, the number of stations is different between the source network and target network, which may cause the dimensions of the source task and target task mismatch. Hence, how many stations are included in a task should be considered before the knowledge transfer is implemented. Second, due to the lack of data, it is difficult to capture adequate long-period spatial-temporal characteristics in the target meta-learning task (i.e., the task belongs to the stations with insufficient data). Therefore, how constructing meta-learning tasks that can borrow sufficient long period spatial-temporal characteristics from source stations to the target task is required.

**(2) Transfer the knowledge from source stations to target stations**

The spatial-temporal characteristics of passenger flows vary from task to task, while the inclusion of multiple stations in each task enhances the diversity of spatial-temporal characteristics. Whether the spatial-temporal distribution features of the target tasks match that of the source tasks would affect the prediction accuracy when we transfer the knowledge, i.e., the high similarity of the passenger flow distribution between the source task and the target task will improve the prediction accuracy. Conversely, low similarity will reduce prediction accuracy. Hence, the transfer-learning based prediction model should effectively capture the spatial-temporal distribution features of passenger flow in target tasks, even if the data is limited in target tasks, i.e., it would be more appropriate for a model to be sensitive and comprehensive enough to various spatial-temporal distribution patterns during the transferring process.

To tackle the challenges, we propose a deep learning network called Meta Long Short-Term Memory network (Meta-LSTM). It is the first to utilize the meta-learning paradigm to address the problem of passenger flow prediction in newly-operated URT stations. We solve the first challenge in three steps: (1) we select URT stations with sufficient data as the source domain (e.g., the Beijing URT network), and determine the size of the task according to the number of target stations. (2) We select the specific stations which are involved in each task according to the relative positional relationship of the stations. (3) The real-time, daily, and weekly patterns are used to capture the long-term temporal patterns of passenger flow in each task.

For the second challenge, we process in two steps: (1) We combine the meta-learning paradigm with a basic network (i.e., LSTM), and train the combined model (i.e., Meta-LSTM) on the source task dataset, which makes it remarkably sensitive to various spatial-temporal scenarios. (2) After training the model in the source stations, the model parameters are extracted and applied as initialization to the passenger flow prediction model of the target stations. Note that the target domain uses the same basic network as the source domain. The evaluation results on the datasets of newly-operated stations and stations in operation for several years show that the performance of our proposed Meta-LSTM is better than several competitive baseline models. The major contributions of this paper are as follows:

(1) A Meta-LSTM model is proposed for predicting the passenger flow in newly-operated URT stations. To the best of our knowledge, this is the first work to conduct the short-term passenger flow prediction for the newly-operated URT stations utilizing meta-learning methods.

(2) We propose a task construction method according to the relative positional relationship of the stations in the Meta-LSTM model. The real-time, daily, and weekly patterns are used to capture the long-term temporal patterns of passenger flow in each task.

(3) Our proposed Meta-LSTM learns the knowledge from existing sufficient passenger flow data in URT stations and transfers the knowledge to data-scarce stations. It can improve the performance when the passenger flow characteristics between two domains do not match. The results show that our model outperforms several competitive baseline models.

The remainder of this paper is organized as follows. In Section 2, related work of this work is reviewed and discussed. Section 3 presents the definitions and problem formulation. In Section 4, we first demonstrate how to construct the meta-learning task. Then we introduce the details of Meta-LSTM. In Section 5, we apply our model to two real-world datasets and discuss the results. In Section 6, the conclusion and future work are discussed.

## 2. Related works

Spatial-temporal prediction is a fundamental task for URT operational management in the era of big data. The conventional time series prediction methods are widely used in the past several years, such as the HA method, ARIMA, Kalman filtering method, etc. Many researchers have also summarized and analyzed these models (Vlahogianni, Golias, and Karlaftis 2004). Machine learning was subsequently developed (P. Wang and Liu 2008b; X. Wang et al. 2018b). In recent years, deep learning methods become prevailing. Ma et al. (2015b) first proposed a LSTM model to predict the traffic speed. After that, many classical deep learning methods are introduced into the traffic spatial-

**7 / 37**

temporal prediction field, such as gated recurrent units (GRU) (Fu, Zhang, and Li 2016), CNN (W. Zhang et al. 2019), ST-ResNet (Junbo Zhang, Zheng, and Qi 2017), and ST-GCN (B. Yu, Yin, and Zhu 2018). After this stage, hybrid models based on RNN, CNN, and GCN emerge. Jinlei Zhang et al. (2021) combines residual architecture with LSTM for short-term passenger flow forecasting in URT. Jinlei Zhang et al. (2020) introduced a model which combines a GCN and a three-dimensional CNN for short-term passenger flow forecasting in URT. Besides, spatial-temporal prediction methods in the URT domain are also applied to various conditions. The methods proposed by(Yang Liu, Liu, and Jia 2019; Jinlei Zhang, Chen, and Shen 2019) were utilized in predicting short-term passenger flow for single or multiple URT stations. Under the condition of normal and abnormal, Jin et al. (2020), R. Yu et al. (2020), W. Li et al. (2021) proposed different models for passenger flow prediction in the traffic field.

Overall, an amount of data needs to be provided if the spatial-temporal models aforementioned perform favorably. However, there is insufficient passenger flow data in newly-operated stations. Spatial-temporal prediction models that are more dependent on sufficient data may not be able to achieve satisfactory performance. Thus, it's necessary to introduce a framework that can not only take advantage of the aforementioned methods to capture spatiotemporal knowledge on multiple stations with sufficient data but also can transfer the knowledge to data scarcity stations.

Meta-learning is an effective means for the challenges of data scarcity in various machine learning applications (Mandal et al. 2021). It leverages prior learning experience to adapt to new tasks quickly, and then learns a useful algorithm for the new tasks with few samples. The taxonomy of meta-learning contains three aspects: meta-representation, meta-optimizer, and meta-objective (Hospedales et al. 2021). Meta-representation shows what meta-knowledge should be chosen for meta-learning. It covers a wide range, such as parameter initialization (Y.-S. Chen, Chiang, and Wu 2022; Song et al. 2022; Yao et al., n.d.), the optimizer of meta-learning (Casgrain and Kratsios 2021), feed-forward models of meta-learning (Triantafillou et al. 2020; Requeima, Gordon, and Bronskill,

n.d.), loss-learning (Bechtle et al. 2021), and so on. The meta-optimizer means how can we optimize a meta-learning model. The gradient descent method occupies a large chunk of all meta-optimizers (Y. Chen et al. 2021; Franceschi et al., n.d.). Last but not least, meta-objective holds a key position in three aspects because the meta-objective represents the goal of meta-learning. In the field of traffic prediction, to improve the prediction accuracy of data-scarcity cities, Yao et al. (2019), Tian et al. (2021) transferred the knowledge of traffic flow from multiple data-rich cities to the cities with limited data, which were achieved by a MAML-based model and a Reptile-based model, respectively. The model proposed by (C. Li et al. 2021) learned correlative demand patterns from bus mode with intensive stations and adapt the knowledge to the station-sparse mode (e.g., train, light rail, and ferry) to boost the prediction performance. Y. Zhang et al. (2022) focused on the prediction of urban traffic speed and volume, they trained multiple traffic prediction tasks segmented by historical traffic data to learn a strategy that can be quickly adapted to related but unseen traffic prediction tasks.

In summary, meta learning framework gives us a strategy for learning knowledge in historical experience and transfer knowledge to new tasks according to requirements. Fortunately, spatial-temporal prediction methods perform a strong capability for capturing spatial-temporal characteristics. Therefore, we construct a Meta-learning model (i.e., Meta-LSTM) for spatial-temporal prediction tasks, which can learn the knowledge from data-rich URT stations and transferring the knowledge to the URT stations with limited data.

## 3. Problem Definitions

In this section, we first define the notations used in this paper and then give the problem definition of short-term passenger flow prediction for the newly-operated subway stations.

**Definition (Spatial-temporal series):** Suppose there are N URT stations. According to the granularity of data collection, we divide the research period into t time intervals. Then the spatial-temporal series of stations i is denoted as a tensor $X^i = [x_1^i, ..., x_t^i]$, where $i \in N$. For a meta-learning task containing I stations, the spatial-temporal series of the task is denoted as $X^I = [X^1, ..., X^I]$.

**Problem formulation:** Given a set of source stations $S_u = \{s_1, ..., s_u\}$ and a set of target stations $S_t = \{s_1, ..., s_t\}$, our goal is to utilize the passenger flow in previous $\tau$ time steps to predict the passenger flow of the $S_t$ in the next time step, which is on the premise that the $S_t$ lacks data (i.e., newly-operated stations) and the $S_s$ provides sufficient data. It can be described as follow:

$$\hat{Y}_{\tau+1} = f\left(X_1^{S_t}, \cdots, X_\tau^{S_t} | f_{\theta_0}^{S_u}\right), \tag{1}$$

where $\hat{Y}_{\tau+1}$ is the passenger flow in time step $\tau + 1$. $f$ represents the function for learning knowledge from source stations and transferring the knowledge to target stations, which will be discussed in detail in Section 4.2. $\theta_0$ denotes the initialization which is extracted from source stations and transferred to target stations.

## 4. Methodology

Herein, we first introduce how to construct the task for learning knowledge. (i.e., meta-learning task). Then, we describe the method we proposed that can learn the knowledge from data-rich stations via multiple meta-learning tasks and transfer the learned knowledge to the data-limited stations (i.e., Meta-LSTM).

### 4.1. Meta-learning task construction

The meta-learning task is a matrix that contains the spatial position of the stations and the spatial-temporal information of passenger flow. We construct all source stations and target stations in the order of URT lines into a one-dimensional sequence, respectively. The one-dimensional vector formed by the stations carries the diverse relative position information of the station (e.g., intermediate station and intermediate station, terminal station and intermediate station, terminal station and terminal station on different lines). We set the size of a meta-learning task, namely the station number in a single meta-learning task, according to the number of target stations. For example, as shown in Fig.1, there are ten stations in the target domain, such that we select the first ten stations from the source domain as the first meta-learning task and follow-up ten stations as the second task. Then the station number of multiple tasks is confirmed accordingly.

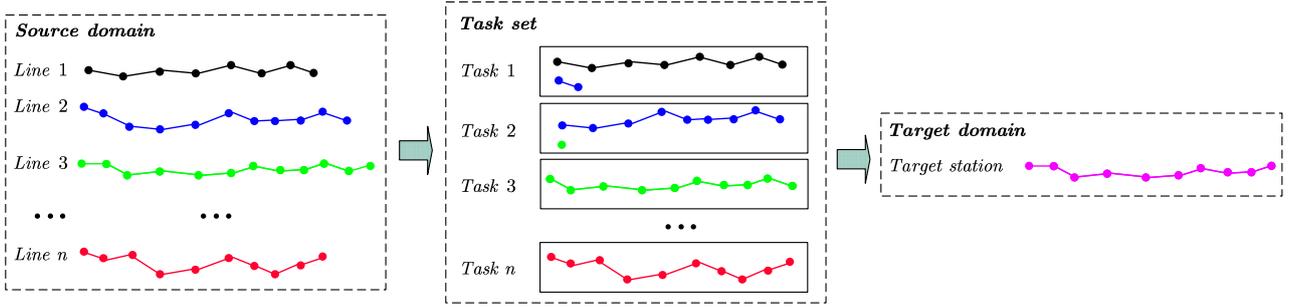

**Fig. 1.** The construction process of a meta-task. There are ten stations in the target domain.

The passenger flow in URT stations has spatial-temporal dependencies. The current passenger flow of a station is affected by recent time intervals. The passenger flow may be similar on consecutive weekdays. Moreover, the same distribution of passenger flow may gradually happen every week. Besides, the passenger flow of different stations may affect each other. Therefore, based on obtaining the location of the stations, we set up diverse passenger flow patterns (i.e., real-time, daily, and weekly patterns) to capture long-period spatial-temporal dependencies (Jinlei Zhang et al. 2021) (shown in Fig.2). The time series of passenger flow in a station is shown as follows:

$$X_t = \begin{bmatrix} (x_{t-\tau}^r \ x_{t-\tau+1}^r \ x_{t-\tau+2}^r \ \cdots \ x_t^r) \\ (x_{t-\tau}^d \ x_{t-\tau+1}^d \ x_{t-\tau+2}^d \ \cdots \ x_t^d) \\ (x_{t-\tau}^w \ x_{t-\tau+1}^w \ x_{t-\tau+2}^w \ \cdots \ x_t^w) \end{bmatrix}, \qquad (2)$$

in which $t$ represents the $t-th$ time interval, and $\tau$ is the number of historical time steps. And $x_t^r, x_t^d, x_t^w$ represent the passenger flow at $t$ time interval in real-time pattern, daily pattern and weekly pattern, respectively. Taking source stations as an example, the time series of multiple stations are defined as Eq. (3), where $u$ is the index of stations.

$$X_t^* = \left[X_{1,t},\ X_{2,t},\ X_{3,t},\cdots,X_{u,t}\right]^T, \tag{3}$$

In summary, the diversity of station location and passenger flow information ensures the diversity of meta-learning tasks, thus further ensuring the stability of the meta-learning method.

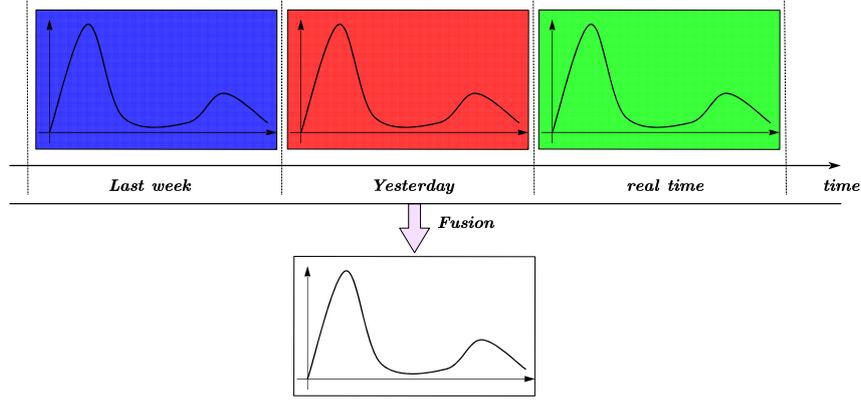

**Fig. 2.** Long-period spatial-temporal dependencies are captured by the fusion of real-time, daily, and weekly patterns.

## *4.2. Knowledge learning and transferring*

To learn knowledge from source stations and transfer the knowledge to target stations, we propose a meta-learning method, which involves three parts: basic network, meta-learner, and applying knowledge to target stations. (1) Basic network. It is utilized to learn the spatial-temporal characteristics of passenger flow. (2) Meta-learner. It is a framework that improves the generalization ability of the basic network to various passenger flow characteristics via multiple meta-learning tasks. (3) Applying the knowledge to target stations. It is to apply the learned knowledge in the form of parameter initialization to passenger flow prediction at target stations. Note that the model used for prediction at target stations is also the basic network aforementioned.

### *4.2.1. Basic network*

The long short-term memory network (LSTM) is utilized as the basic network. The LSTM can capture the long short-term temporal information, which is effective in predicting passenger flow. LSTM has a chained form with a duplicate neural network module which has four layers interacting extraordinarily. The repeating module named the memory unit consists of the cell status, an input gate, a forget gate, and an output gate (as shown in Fig.3). Cell status structure provides a path for messaging, and the gate structures enable cell status to overcome long-short term memory problems by selectively remembering or forgetting information.

The forget gate controls how much information is obtained from the cell status at the previous moment, which is determined by the value of the forgetting gate, as shown in Eq. (4).

$$\Gamma_f^t = \sigma(W_f[h^{t-1}, x^t] + b_f), \qquad (4)$$

where $\Gamma_f^t$ denotes the current forget gate, σ is an activation function named sigmoid function. *W* and *b* are learnable weight and bias tensors $h^{t-1}$ named hidden state is the information from the previous sequence, $x^t$ is the input data at present. Note that the definition of $W$ and $b$ in the rest of this section is the same as here, if not specified.

The input gate $\Gamma_i$ determines how much information can be delivered from the current moment, which uses an activation function σ to convert selected information into a form that can be added to the cell status. Then it prepares a candidate cell status vector $\tilde{c}^t$ for updating cell status. The process is shown as follows:

$$\Gamma_i^t = \sigma(W_i[h^{t-1}, x^t] + b_i), \qquad (5)$$

$$\tilde{c}^t = tanh(W_c[h^{t-1}, x^t] + b_c). \qquad (6)$$

Finally, we combine forget gate and input gate to update cell status $c^t$ at the current moment, and set an output gate $\Gamma_o$ to control the output information.

$$c^t = \Gamma_f^t \cdot c^{t-1} + \Gamma_i^t \cdot \tilde{c}^t, \tag{7}$$

$$\Gamma_o^t = \sigma(W_o[h^{t-1}, x^t] + b_o), \tag{8}$$

$$h^t = \Gamma_o^t \cdot tanh(c^t). \tag{9}$$

The value range of the three gates mentioned above is all in $[0, 1]$. Note that the memory unit saves no information or all information if the value of the gate is 0 or 1, and other values in $(0, 1)$ means only partial information could deliver to the next part. Subsequently, $h^t$ encodes the passenger flow information of a certain station. The passenger flow sequence of the next time step can be predicted by

$$y_{pred}^t = tanh(W_y \cdot h^t + b_y). \tag{10}$$

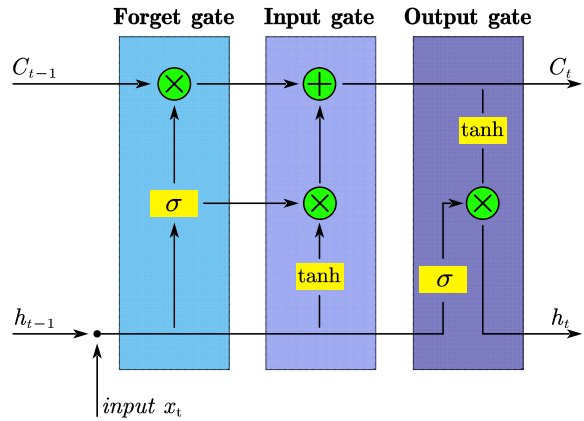

**Fig.3.** Overview of the LSTM which is utilized as the basic network.

*4.2.2. Meta learner*

Meta-learner is a framework that increases the generalization ability of the basic network to various passenger flow characteristics via meta-learning tasks. The meta-learning task set is constructed as Fig. 4. As suggested in (Finn, Abbeel, and Levine 2017), to increase the generalization ability of the basic network to various passenger flow characteristics, we aim to optimize the model parameters with a small number of gradient steps on each task from source stations and then minimize the average of generalization losses of overall source stations (shown as Fig. 5), such that the performance is effective on a new task although a little data is provided. The optimal object of meta learner is presented as follows:

$$\theta_0 = arg \min_{\theta_i^*} \sum_{T_i \sim p(T_s)} L_{T_i}(\hat{Y}_{T_i}(\theta_i^*), Y_{T_i}) \tag{11}$$

in which $\theta_0$ is the parameter that will be transferred to target stations (i.e., the optimal weights and bias). We consider a distribution over tasks $p(T_s)$ that we draw from source stations. The task $T_i$ is sampled from $p(T_s)$. The parameter $\theta_i^*$ is outputted by the basic network when the train task set is applied. Note that $\theta_i^*$ does not necessarily minimize the loss of single task in the train task set while it is necessary to minimize the total loss of all tasks in the test task set. $\hat{Y}_{T_i}(\theta_i^*)$ represents the predicted value for task $T_i$ when the parameter is $\theta_i^*$. $Y_{T_i}$ is the true value for task $T_i$. $L$ is a loss function.

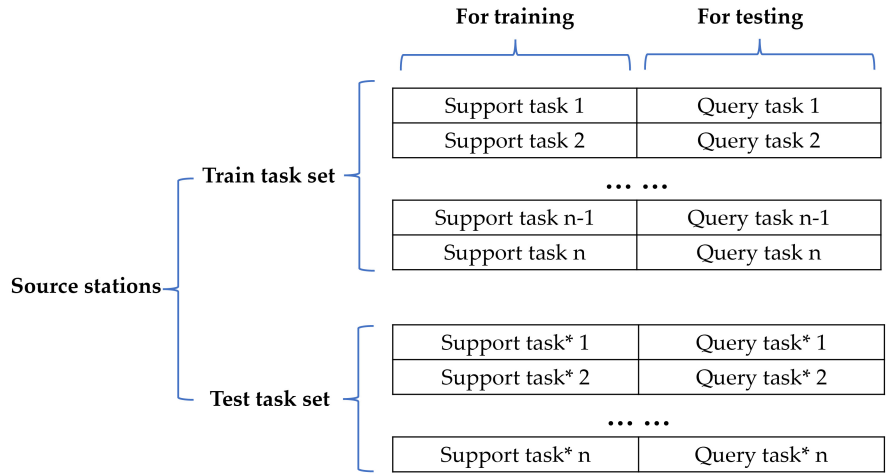

**Fig. 4.** Structure of meta-learning task set in source stations. It contains two parts, i.e., train task set and test task set. In each task set, the support task is used for training and the query task is used for testing.

The meta-learner consists of local-learner and global-learner, which is shown in Fig.5. The local learner is utilized to improve the ability of the basic network to capture the spatial-temporal characteristics of passenger flow in a single meta-learning task. The local-learner is trained with meta-learning task $T_i$ (i.e., the task in the train task set) and feedback from corresponding $loss$, and then tested on new tasks which are sampled from the test task set. One local learner is taken as an example which is shown in Fig. 6. For further explanation, suppose there is a parametrized function $f_\theta$ with parameters $\theta$. We update the parameters $\theta$ to $\theta_i^*$ when we adapt to a new task $T_i$ (shown as Fig.7 a). Taking one gradient update as an example, the local learner is as follows:

$$\theta_i^* = \theta - \alpha \nabla_\theta L_{T_i}(f_\theta), \tag{12}$$

in which $\alpha$ is the learning rate of the local-learner, and $\nabla$ represents the operation of gradient descent.

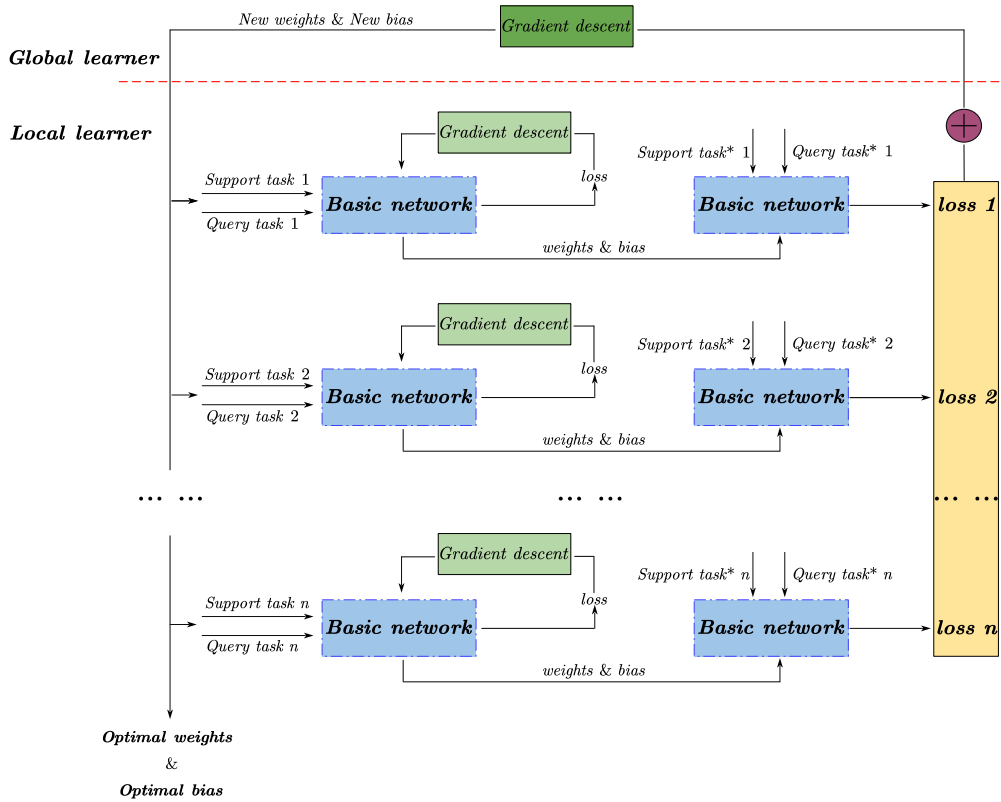

**Fig.5** Overview of the meta-learner. The part above the red dotted line is the global learner, and the part below the red dotted line is the local learner.

The global-learner enables the basic network to obtain superior generalization performance to various tasks. The updating process of the global-learner is by optimizing the performance of $f_{\theta_i^*}$ with respect to $\theta$ across all tasks sampled from $p(T_s)$ (shown as Fig.7 b). The learning process is presented in the form of parameter iterative updates, i.e.,

$$\min_{\theta} \sum_{T_i \sim p(T_s)} L_{T_i}(f_{\theta_i^*}) = \sum_{T_i \sim p(T_s)} L_{T_i}\left(f_\theta - \alpha \nabla_\theta L_{T_i}(f_\theta)\right), \tag{13}$$

$$\theta \leftarrow \theta - \beta \nabla_\theta \sum_{T_i \sim p(T_s)} L_{T_i}(f_{\theta_i^*}), \tag{14}$$

where $\beta$ is the learning rate of the global-learner.

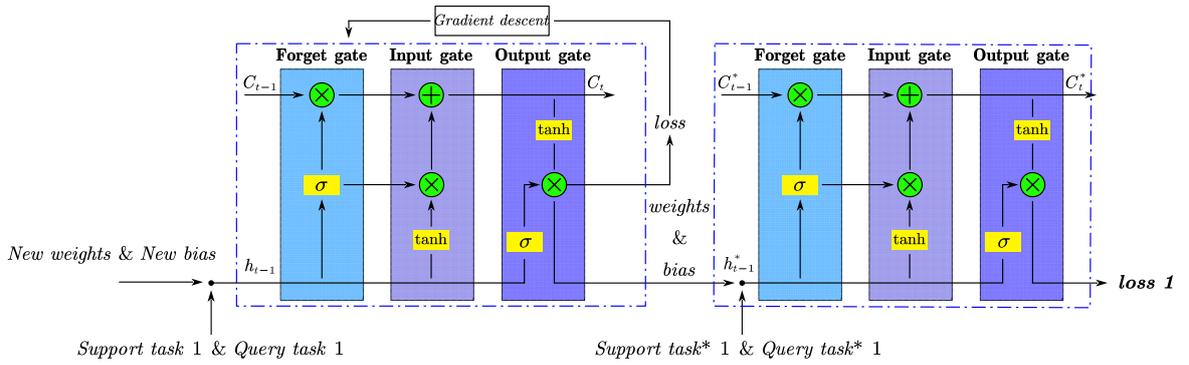

**Fig.6** the structure of a local-learner. The structure within the blue dotted frame is the basic network. New weights and new bias are provided by global-learner.

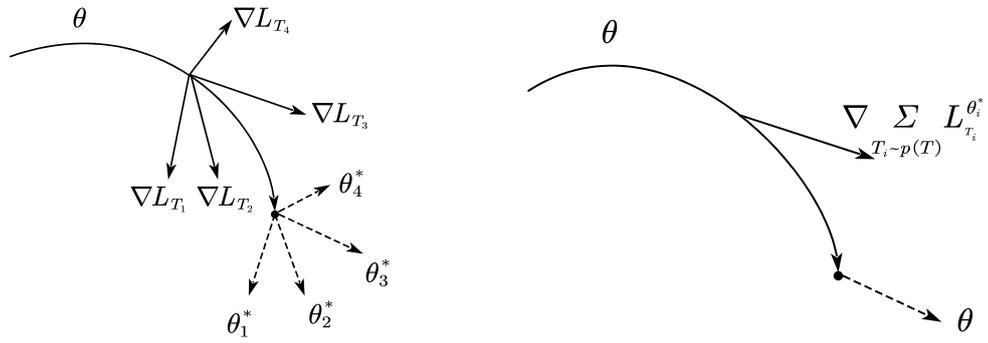

**a.** Updating process of local learner.    **b.** Updating process of global learner.

**Fig. 7.** Updating process of local learner and global learner.

We expand the gradient part on the right of Eq. (14), and we can get:

$$\nabla_\theta \sum_{T_i \sim p(T)} L_{T_i}(f_{\theta_i^*}) = \sum_{T_i \sim p(T)} \frac{\partial L_{T_i}(f_{\theta_i^*})}{\partial \theta_i^*} \frac{\partial \theta_i^*}{\partial \theta}. \tag{15}$$

According to Eq. (12), the Eq. (15) can be expressed as follows:

$$\sum_{T_i \sim p(T)} \frac{\partial L_{T_i}(f_{\theta_i^*})}{\partial \theta_i^*} \frac{\partial \theta_i^*}{\partial \theta} = \sum_{T_i \sim p(T)} \frac{\partial L_{T_i}(f_{\theta_i^*})}{\partial \theta_i^*} \left[1 - \alpha \cdot \partial \left(\frac{\partial L_{T_i}(f_\theta)}{\partial \theta}\right) / \partial \theta \right]$$
$$= \sum_{T_i \sim p(T)} \frac{\partial L_{T_i}(f_{\theta_i^*})}{\partial \theta_i^*} \left[1 - \alpha \cdot \left(\frac{\partial^2 L_{T_i}(f_\theta)}{\partial \theta^2}\right)\right], \tag{16}$$

in which exists a second-order differential that would increase the computational complexity, and drop the computational efficiency. Hence, it is necessary to lower the order of differentiation by the following simplification. In URT, passenger flow prediction is a multiple linear regression problem. Hence, the loss function of local learner and global learner are both multiple linear regression problems, which signifies the second-order partial derivatives of the loss function are zero. i.e.,

$$\left(\frac{\partial^2 L_{T_i}(f_\theta)}{\partial \theta^2}\right) = 0 \tag{17}$$

Then the updating process for global-learner would be simplified as follow:

$$\theta \leftarrow \theta - \beta \nabla_{\theta_i^*} \sum_{T_i \sim p(T)} L_{T_i}(f_{\theta_i^*}). \tag{18}$$

After the proper simplification, only the first-order gradient is considered in the process of parameter update, which also simplifies the calculation process. Then we can transfer the initialization $\theta_0$ to target stations. The target stations can learn new knowledge based on prior knowledge, and obtain superior generalization performance. The details of transferring process are described in the following.

*4.2.3. Apply knowledge to target stations*

To improve the prediction performance in target stations, we first utilize the prior knowledge $\theta_0$ as initialization parameters for pre-training in target stations, and then we apply the weights and bias obtained by pre-training for testing (shown as Fig.8). The parameters are updated by iterative updating. One gradient update process is taken as an example:

$$\theta_{t_n} = \theta_0 - \gamma \nabla_{\theta_0} L_{tn}(f_{\theta_0}), \tag{22}$$

where $L_{tn}$ is the loss function in target stations.

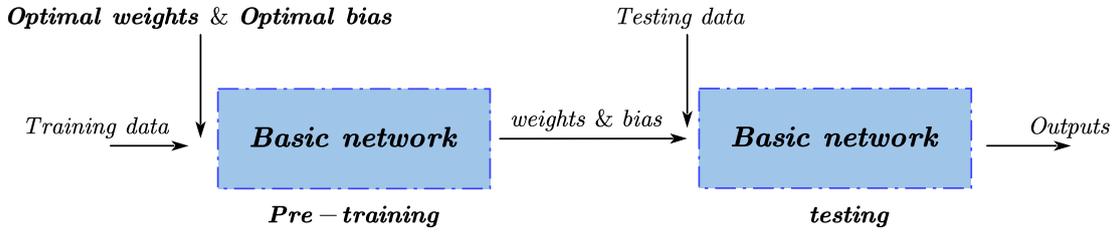

**Fig.8** The process of applying knowledge to target stations.

## 5. Experiments

In what follows we evaluate the proposed meta-learning method based on several URT datasets. We first describe the datasets and model configuration. Then we compare the performance of our proposed Meta-LSTM with baseline models and discuss the results.

*5.1. Dataset description*

We collect three AFC datasets of URT from three cities: 1) Beijing subway dataset (05:00 - 23:00, 2016-02-29 to 2016-04-03), 2) Hangzhou subway dataset (06:00 - 23:00, 2019-01-01 to 2019-01-25) and 3) Nanning subway dataset (06:30 - 22:00, 2016-06-28 to 2016-07-17). In the Beijing subway dataset, there are 17 lines with 276 subway stations (excluding the airport express line and the corresponding stations) in March 2016. The 17 lines of the Beijing subway were operated between 1971 and 2014. In the Nanning subway dataset, there is 1 line with 10 stations newly-operated in June 2016. In the Hangzhou subway dataset, there are 3 lines with 80 subway stations in January 2019. The 3 lines of the Hangzhou subway were operated between 2012 and 2015. The detailed information on subway lines in operation is listed in Table 1. Note that only data of workdays are considered. The time interval of passenger flow is 15 minutes.

**Table 1** Duration of operation of subway lines.

| City | Timespan (m/d/y) | Line | Operational time | Duration of operation (years) |
|---|---|---|---|---|
| Beijing | 2016/02/29 – 2016/04/03 | 1 | 1971-01 | 45 |
| | | 2 | 1971-01 | 45 |
| | | 13 | 2002-09 | 13 ~ 14 |
| | | Ba Tong | 2003-12 | 12 |
| | | 5 | 2007-10 | 8 ~ 9 |
| | | 10 | 2008-07 | 7 ~ 8 |
| | | 8 | 2008-07 | 7 ~ 8 |
| | | 4 | 2009-09 | 6 ~ 7 |
| | | 15 | 2010-12 | 5 |
| | | Chang Ping | 2010-12 | 5 |
| | | Fang Shan | 2010-12 | 5 |
| | | Yi Zhuang | 2010-12 | 5 |
| | | 9 | 2011-12 | 4 |
| | | 6 | 2012-12 | 3 |
| | | 14 | 2013-05 | 2 ~ 3 |
| | | 7 | 2014-12 | 1 |
| Hangzhou | 2019/01/01 – 2019/01/25 | 1 | 2012-11 | 6 ~ 7 |
| | | 2 | 2014-11 | 4 ~ 5 |
| | | 4 | 2015-02 | 3 ~ 4 |
| Nanning | 2016/06/28 – 2016/07/17 | 1 | 2016-06 | Newly-operated |

Compared with Beijing URT, the characteristics of passenger flow distribution are insufficient, although the three lines of Hangzhou URT have been in operation for some time. The Nanning URT stations are all newly operated. Hence, we use Beijing subway stations as the source stations and the stations of Nanning subway and Hangzhou subway as the target stations. For each source station, we select 80% of the data for training and validation, and the rest for testing. For each target station, we select the 1-day, 3-day, and 5-day data for training (i.e., the amount of data used for training was 1 day, 3 days, 5 days), and the rest for testing.

*5.2. Model Configuration*

During the training procedure of the source stations, each LSTM layer consists of 32 neural units. The inputs and outputs of LSTM depend on the number of stations in each meta-learning task. We set the learning rate of local learners and global learner as 0.001. The number of updates for each meta-task is set as 5. We use Adam as the optimizer. The training batch size for each meta-iteration is set as 16, and the maximum iteration of meta-learning is set as 40000. During the training procedure of the target stations, we set the learning rate as 0.01. The training batch size for each iteration is set as 16. We also use the Adam optimizer to train the model.

We use the mean-squared error (MSE) as the loss function. To balance the model training time and prediction performance, for each passenger flow pattern, we use the previous five timesteps to forecast the next one by trial and error. In the training process, we use the model checkpoint and early stopping technique to save the best model and avoid overfitting.

*5.3. Baseline models*

We set two kinds of baseline models that include non-transfer models and transfer models. For the non-transfer models, we use the training data of target stations for training and use the rest for evaluation, such as ARIMA, HA, LSTM, CNN, and ST-ResNet. For the transfer models, we use the training data of source stations to train them and apply the knowledge to target stations, such as Fine-tuning Method and Meta-CNN. The details of baseline models are as follows.

**ARIMA:** A representative mathematical statistics-based model for time-series prediction.

**HA:** HA model makes a prediction using the average of historical passenger flow data. For example, we use the HA model to calculate the mean value of the passenger flow of all time from 8:00 am – 9:00 am, and then we use the result as the predictions at 8:00 am – 9:00 am in current.

**LSTM:** It is a Long Short-Term Memory network. The details of the LSTM are covered in Section 4.2.1. Specifically, the LSTM model is established with one hidden layer and two fully connected layers. Each LSTM layer consists of 32 neural units. The optimizer is Adam with a learning rate of 0.01. The inputs and outputs of LSTM depend on the number of stations in each meta-learning task.

**CNN:** Convolutional Neural Networks (CNN) is a kind of feedforward neural network with deep structure and convolution computation. We set four convolution layers and one fully connected layer. The parameters of the CNN layer are out channels = 1, kernel size =3, stride = 1 and padding = 1, respectively. The optimizer is Adam with a learning rate of 0.01. The inputs and outputs are the same as LSTM.

**Spatial-Temporal Residual network (ST-ResNet):** This method considers the spatial-temporal relationship of passenger flow data based on the residual network. We set two residual blocks, and each block consists of two convolution layers. The parameters of the convolution layer are the same as CNN.

**Fine-tuning Method (FT):** This method is to select a meta-learning task in source stations randomly and then train a basic network in the selected meta-learning task, and finally fine-tune the basic network for target stations. We set two kinds of Fine-tuning methods:

**(1) FT-CNN:** The CNN is selected as the basic network. The parameters of the convolution layers are the same as CNN.

**(2) FT-LSTM:** The LSTM is selected as the basic network. The parameters are the same as LSTM above mentioned.

**Meta-CNN:** This method uses the same framework as Meta-LSTM, while the basic network is replaced by CNN. The parameters of the meta-learner are the same as Meta-LSTM, and the parameters of the convolution layer are the same as CNN.

*5.4. Performance criteria*

We select root mean squared error (RMSE), mean absolute error (MAE), and weighted mean absolute percentage error (WMAPE) as evaluation indicators to evaluate model performance. The indicators are given by Equations (23) - (25):

$$RMSE = \sqrt{\frac{1}{n}\sum_{i=1}^{n}(y_i - \hat{y}_i)^2}, \tag{23}$$

$$MAE = \frac{1}{n}\sum_{i=1}^{n}|(y_i - \hat{y}_i)|, \tag{24}$$

$$WMAPE = \sum_{i=1}^{n}\left(\frac{y_i}{\sum_{j=1}^{n}y_j}\left|\frac{y_i - \hat{y}_i}{y_i}\right|\right), \tag{25}$$

where $\sum_{j=1}^{n}y_j$ is the sum of actual values.

*5.5. Results and discussion*

In this section, we evaluate our model and compare it with baseline models on two subway datasets. The results involve three parts: 1) Passenger flow prediction for newly-operated stations (i.e., Nanning subway stations). The prediction performances are shown in Table 2 and Fig. 9. 2) Passenger flow prediction for stations that have been in operation for several years (i.e., Hangzhou subway stations), and the prediction performances are shown in Table 3 and Fig. 10. 3) Analyze the improvement of the meta-learning framework on predictive performance. For ease of writing, we define CNN, FT-CNN, Meta-CNN, and ST-ResNet as CNN-based models, and define LSTM, FT-LSTM, and Meta-LSTM as LSTM-based models.

*5.5.1. Prediction performances of the newly-operated stations*

From the aspect of model categories, the performance of ARIMA and HA are worse than CNN-based models while the CNN-based models perform worse than LSTM-based models. On the experiments with 1-day data, the performance of ARIMA is worse than CNN-based models and LSTM-based models. The ARIMA model almost fails to give favorable prediction results. On the experiments with 3-day and 5-day data, the performance of ARIMA has significantly improved while it is still worse than other models. The results indicate that the passenger flow of the newly-operated stations is instability with poor regularity. As for the ARIMA model, it predicts current passenger flow based on historical values and prediction errors in historical values. The prediction errors of the model would be decreased and we could get the perfect performance of prediction if the data possesses a strong daily periodicity (Yao et al. 2019). However, the daily periodicity of passenger flow in newly-operated stations has not been established. It is difficult for ARIMA to find the daily periodicity, which decreases the accuracy of ARIMA. With the increase in data volume, the regularity of passenger flow appears gradually, and the performance is improved. On the experiments with 1-day data, the performance of HA is worse than CNN-based models and LSTM-based models. On the experiments with 3-day and 5-day data, the performance of HA has been improved. The HA model outperforms almost all CNN-based models (except for the performance of ST-ResNet under experiments with 5-day data) while it is still worse than all LSTM-based models. The results indicate that, unlike the ARIMA, the passenger flow prediction by the HA model is based on the average values of historical data, which naturally reduces the impact of passenger flow instability on prediction results. However, the passenger flow information has also been reduced when the HA model uses the average values of historical data for prediction. Compared with the HA model, there is a better performance contributed by LSTM-based models, especially our proposed Meta-LSTM which aims to capture all characteristics of historical passenger flow.

**Table 2** Performance comparison for passenger flow prediction on Nanning subway dataset.

| Passenger flow | Nanning subway line 1 | | | | | | | | |
| --- | --- | --- | --- | --- | --- | --- | --- | --- | --- |
| | 1-day | | | 3-day | | | 5-day | | |
| | RMSE | MAE | WMAPE | RMSE | MAE | WMAPE | RMSE | MAE | WMAPE |
| ARIMA | 80.13 | 54.13 | 0.999 | 63.245 | 36.19 | 0.633 | 36.89 | 17.93 | 0.311 |
| HA | 60.35 | 35.30 | 0.62 | 29.20 | 17.23 | 0.297 | 24.53 | 14.34 | 0.252 |
| FT-CNN | 42.748 | 29.462 | 0.495 | 31.181 | 19.973 | 0.320 | 26.261 | 17.175 | 0.284 |
| CNN | 42.302 | 30.284 | 0.492 | 30.240 | 19.382 | 0.311 | 24.535 | 16.503 | 0.272 |
| Meta-CNN | 37.880 | 25.288 | 0.418 | 29.005 | 19.027 | 0.304 | 25.100 | 16.460 | 0.270 |
| ST-ResNet | 32.210 | 19.932 | 0.331 | 29.513 | 18.624 | 0.300 | 24.887 | 14.494 | 0.239 |
| FT-LSTM | 29.889 | 17.840 | 0.292 | 26.880 | 15.383 | 0.258 | 23.527 | 13.315 | 0.223 |
| LSTM | 28.638 | 16.698 | 0.280 | 28.134 | 15.351 | 0.252 | 22.312 | 13.181 | 0.213 |
| **Meta-LSTM** | **26.76** | **15.42** | **0.256** | **24.13** | **13.73** | **0.226** | **21.903** | **12.398** | **0.201** |

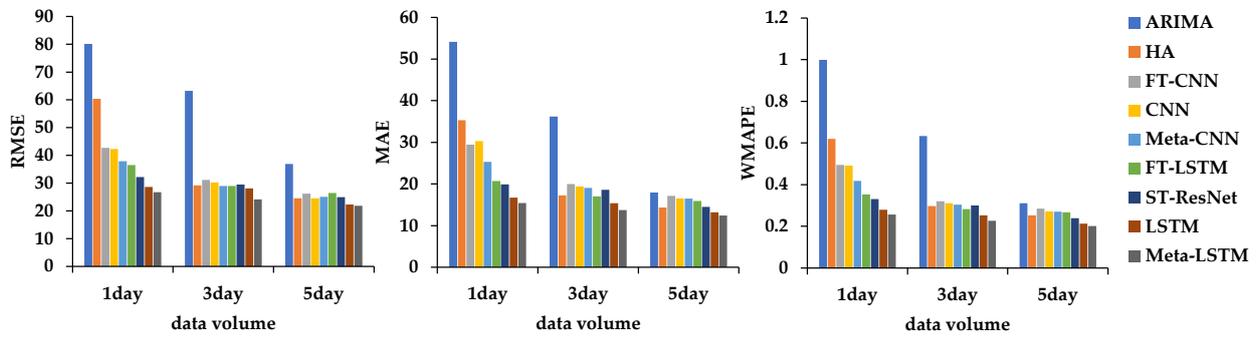

**Fig. 9.** Comparison of Prediction Performances Obtained for Different Models and Data Volume on Nanning Subway Dataset.

On the experiments with 1-day, 3-day, and 5-day data, the performances of CNN-based models are worse than LSTM-based models. The best performance of all models based on CNN is the ST-ResNet. The prediction performance of the other three models (i.e., FT-CNN, CNN, and Meta-CNN) is from poor to good. The results indicate that applying the transfer learning method inflexibly would not necessarily improve the performance of the passenger flow prediction of target stations. When the spatial-temporal distribution of passenger flow is significantly different between the source stations and the target stations (e.g., source stations are from Beijing subway and target stations are from Nanning subway), the inflexible application of the transfer learning method shows poor performance (i.e., the performance of FT-CNN is worse than that of CNN). We use CNN as the basic network of our proposed meta-learning method, which is named Meta-CNN. When we apply the knowledge learning from source stations to target stations by Meta-CNN, we get better performance than CNN. Compared with the other CNN-based models, ST-ResNet considers the temporal and spatial correlations, which is one possible reason that the performance of ST-ResNet is better than other CNN-based models.

The performance of LSTM-based models outperforms the other methods. And our proposed Meta-LSTM achieves the best performance. The performance of FT-LSTM is unstable, and the reason is that spatial-temporal distribution between source stations and target stations is different. It will decrease the accuracy of prediction when the spatial-temporal distribution of source stations and target stations mismatch. Compared with the LSTM, our model not only learns a better initialization but also learns diverse meta-knowledge which consists of long short-term passenger flow information and the information of spatial relative position from meta-tasks of source stations. Then we transfer the knowledge to target stations, which increases the prediction performance.

*5.5.2. Prediction performances of the stations operated for years*

We promote the Meta-LSTM to the stations operated for several years (i.e., the Hangzhou subway network), and analyze the performance in this part. From the perspective of model categories, the performance of conventional prediction models (i.e., ARIMA and HA) is worse than CNN-based models while CNN-based models perform worse than LSTM-based models. The prediction performance sorting of models is similar to that of newly-operated stations.

On the experiments with 1-day data, the performances of ARIMA and HA are worse than other models, and the performances of ARIMA and HA are not much different. On the experiments with 3-day and 5-day data, although the performances of ARIMA and HA have been improved, the accuracy gap between them has increased. The performance of the HA model outperforms all CNN-based models and performs worse than LSTM-based models. The predicted performance of stations operated for years has similar results to that of newly-operated stations (detailed in Table 3). The results indicate that there is a certain periodicity of passenger flow in Hangzhou subway stations (i.e., the stations operated for years) while the passenger volume is still unstable, there is a little information reduced by HA model, because there is a low degree of instability of the passenger flow in Hangzhou subway stations. However, the volume of the HA model to capture the spatial-temporal features is less than LSTM based model, which causes the performance of HA to be worse than LSTM-based models. The ability of ARIMA to capture the passenger flow features is weaker than LSTM-based models and CNN-based models, although we have increased the data volume.

On the experiments with 1-day data, all CNN-based models outperform ARIMA and HA (i.e., sorted by the performance with 1-day data: CNN > HA > ARIMA). On the experiments with 3-day and 5-day data, the performance of CNN based models performs worse than the HA model, and on the experiments with 1-day, 3-day, and 5-day data, all CNN based models perform worse than LSTM based models (i.e., sorted by the performance with 3-day and 5-day data: HA > CNN based model; with 1-day, 3-day, 5-day data: LSTM based models > CNN based models). In all CNN-based models, ST-ResNet contributes the best performance. The performance of Meta-CNN and FT-CNN is the second only to ST-ResNet. The performance of CNN is the worst (i.e., sorted by the performance for CNN-based models: ST-ResNet > Meta-CNN and FT-CNN > CNN). FT-CNN with 1-day and 3-day data outperforms Meta-CNN while the performance of FT-CNN with 5-day data is worse than Meta-CNN. The results demonstrate that simply transferring the knowledge between single objectives is unstable. Through learning knowledge from multiple source stations, the performance of Meta-CNN is more stable than FT-CNN. For the performance of ST-ResNet, we consider the temporal and spatial correlations, which improve the performance significantly.

LSTM-based models perform better than all other models under all experiment data volumes. Our proposed Meta-LSTM achieves the best performance. The FT-LSTM is the second only to Meta-LSTM and the performance of LSTM is the worst in all LSTM-based models (i.e., sorted by the performance for LSTM-based models: Meta-LSTM > FT-LSTM > LSTM). The results suggest that with the increase in station operation time, the passenger flow has a periodicity, such that the performance of the FT method is higher than that of the non-transfer learning method when the passenger flow data is limited. Our proposed Meta-LSTM can learn multiple long-term knowledge, which improves the stability of the results and achieves better performance, although the distribution of passenger flow between source stations and target stations is different.

**Table 3** Comparing with Baselines for Passenger Flow Prediction on Hangzhou Subway Dataset.

| passenger flow | Hangzhou subway line 1 & line 2 & line 4 | | | | | | | | |
|---|---|---|---|---|---|---|---|---|---|
| | 1-day | | | 3-day | | | 5-day | | |
| | RMSE | MAE | WMAPE | RMSE | MAE | WMAPE | RMSE | MAE | WMAPE |
| ARIMA | 280.12 | 128.37 | 0.603 | 151.94 | 73.43 | 0.335 | 112.92 | 68.58 | 0.314 |
| HA | 267.64 | 122.94 | 0.566 | 90.11 | 45.28 | 0.207 | 63.99 | 33.08 | 0.151 |
| CNN | 164.609 | 118.784 | 0.530 | 91.393 | 62.457 | 0.277 | 77.621 | 55.886 | 0.246 |
| FT-CNN | 146.347 | 102.357 | 0.457 | 84.134 | 57.834 | 0.256 | 83.711 | 60.063 | 0.264 |
| Meta-CNN | 145.739 | 104.425 | 0.464 | 92.334 | 60.453 | 0.268 | 71.604 | 52.374 | 0.230 |
| ST-ResNet | 130.151 | 42.796 | 0.411 | 64.487 | 57.972 | 0.187 | 61.566 | 42.409 | 0.186 |
| LSTM | 65.777 | 41.401 | 0.159 | 60.682 | 37.759 | 0.143 | 53.447 | 35.064 | 0.133 |
| FT-LSTM | 65.329 | 40.765 | 0.155 | 59.219 | 36.590 | 0.138 | 53.675 | 34.488 | 0.131 |
| **Meta-LSTM** | **63.864** | **38.531** | **0.149** | **58.700** | **35.828** | **0.136** | **49.771** | **31.585** | **0.120** |

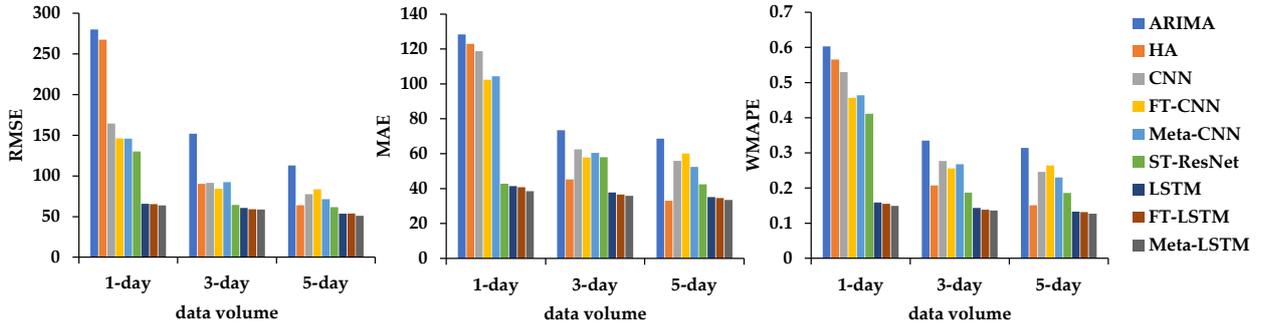

**Fig. 10.** Comparison of Prediction Performances Obtained for Different Models and Data Volume on Hangzhou Subway Dataset.

*5.5.3. The improvement of meta-learning framework on predictive performance*

The prediction performance of Meta-LSTM is the average value of all target stations. How the meta-learning framework improves the performance of the basic network is not intuitive. Hence, we continue to analyze in this part what aspects of the basic network are improved by the meta-learning framework.

We select three typical stations as instantiations. The first one is Huizhanzhongxin station from Line 1 of the Nanning subway. It is located at the intersection of five roads, and it is planned to be a transfer station for Subway Line 1 and Subway Line 4 in the future. It is expected that a large passenger flow enters the station every day. The other two stations, Qianjianglu station, and Fengqi station, are transfer stations of Hangzhou Subway Line 1 and Line 2, Line 2 and Line 4. Huizhanzhongxin station is a newly-operated station. Most passengers enter the station with a tentative attitude, and the daily passenger flow shows a gradual increase (as shown in Fig. 11). The prediction performance of passenger flow in peak hour is better than that of passenger flow in off-peak hours using LSTM. There is little difference between the performance of passenger flow in peak hour predicted by Meta-LSTM and LSTM, and the Meta-LSTM improves the prediction performance of passenger flow in off-peak hours. However, the improvement is insignificant with the increase in data volume.

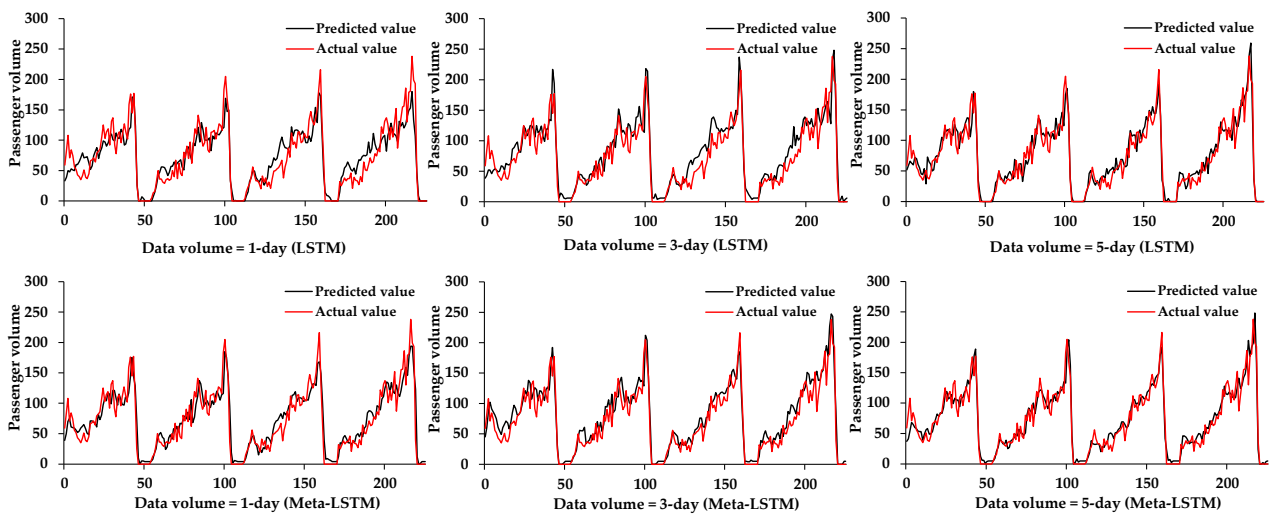

**Fig.11.** Comparison of actual values and predicted values for Huizhanzhongxin station in Nanning subway.

Qianjianglu station and Fengqilu station are located around a largely residential community with thousands of people commuting in the morning, which causes the passenger volume in the morning peak to be significantly higher than that in the evening peak (shown in Fig. 12 and Fig.13).

In both two stations, the prediction performances of passenger flow predicted by Meta-LSTM and LSTM in peak hour are almost consistent. In off-peak hours, Meta-LSTM outperforms LSTM. With the increase of the data volume, the gap between the performance of the two models is gradually narrowed.

In general, for the prediction performance of passenger flow in peak hour, Meta-LSTM and LSTM are indistinguishable. The improvement of meta-learning framework on predictive performance is mainly reflected in off-peak hours, and the affection for improvement becomes insignificant as the data volume is increased. It is one possible reason that the characteristic of passenger flow in peak hour is single (i.e., passenger volume in peak hour is a single peak), such that both models can cope well. The fluctuation of passenger flow in off-peak hours complicates the passenger flow characteristics. Hence, the performance of Meta-LSTM with well generalization ability is better than LSTM.

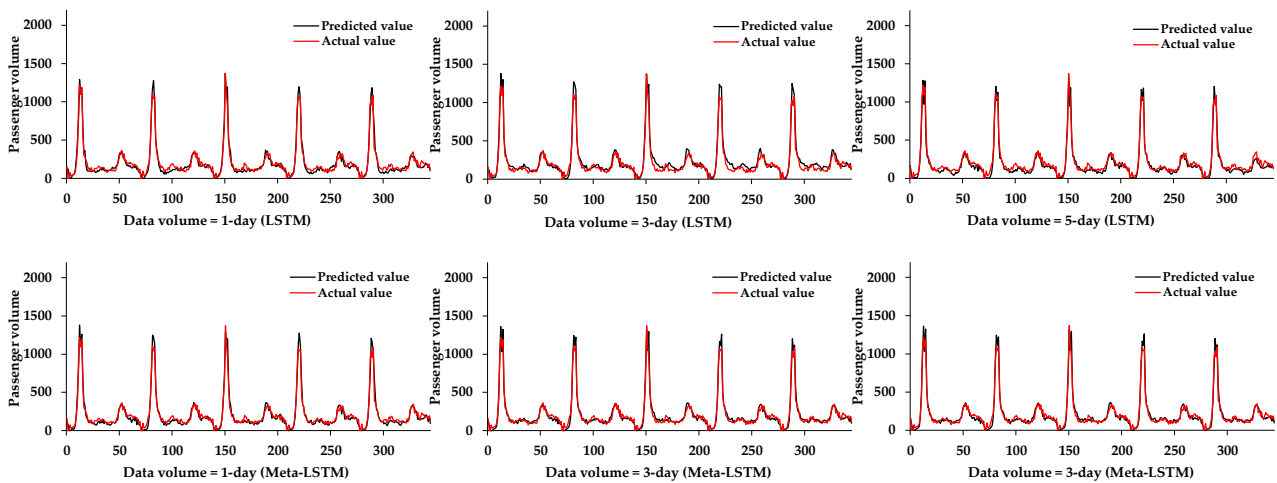

**Fig.12.** Comparison of actual values and predicted values for Qianjianglu station in Hangzhou subway.

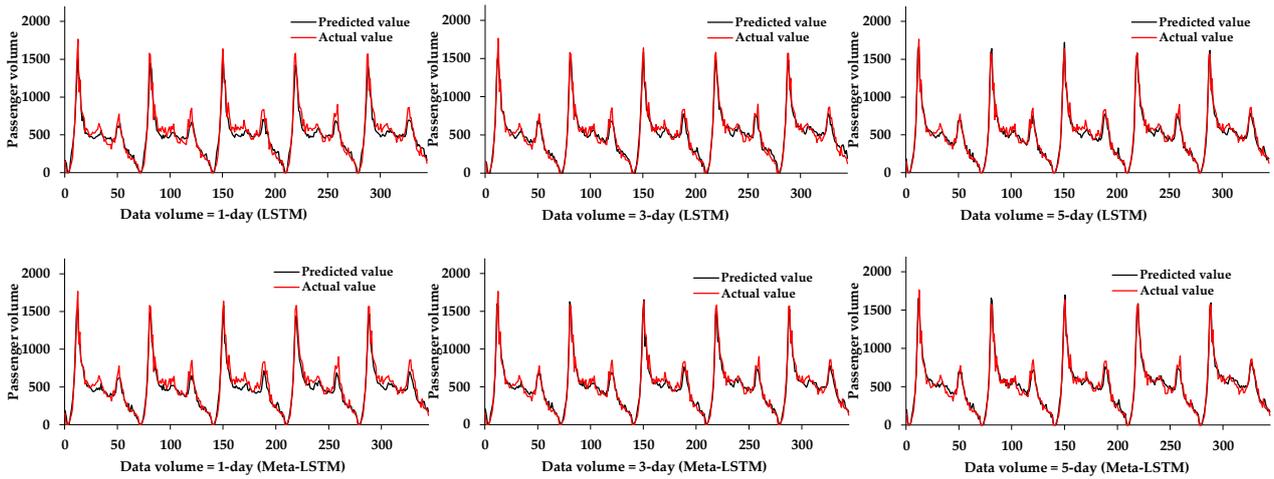

**Fig.13.** Comparison of actual values and predicted values for Fengqi station in Hangzhou subway.

## 6. Conclusion

In this study, we consider the problem of passenger flow prediction with insufficient data over a short-term horizon. Motivated by its application in the operation management for newly-operated stations, we propose a novel Meta-LSTM model which extracts knowledge from the stations with sufficient data and then transfers the knowledge to target stations with limited data in the form of parameter initialization. We test our model on two real-world datasets, including newly-operated stations of the Nanning subway and stations operated for several years of the Hangzhou subway. The main conclusions are summarized as follows.

1. There is a good generalization ability of our proposed Meta-LSTM to various passenger flow characteristics, such that the prediction performance of Meta-LSTM in newly-operated stations (i.e., Nanning subway stations) outperforms several competitive baseline models.

2. We promote the Meta-LSTM to the stations operated for several years (i.e., the Hangzhou subway network) and our proposed Meta-LSTM still performs well when the passenger flow data is limited, which indicates that the Meta-LSTM could perform well in data-scarce stations whether the stations are newly operated or not.

3. The prediction performance is mainly improved in non-peak hours by Meta-LSTM while the performance of Meta-LSTM is similar to LSTM in peak hour, which further corroborates the good generalization ability of Meta-LSTM to various passenger flow characteristics.

In the future, it is potential to make further improvements to the structure of Meta-LSTM by considering the structure of the URT network and fusing more interpretable information (e.g., weather conditions and point of interest) to make the model more comprehensive.

## 7. Conflicts of Interest

The authors declare no conflict of interest.

## 8. Acknowledgments

We wish to thank the anonymous reviewers for the valuable comments, suggestions, and discussions. This work was supported by the National Natural Science Foundation of China (Nos. 72201029, 71825004).